# Opinion Mining on Offshore Wind Energy for Environmental Engineering

Isabele Bittencourt[1,4], Aparna S. Varde[2,4], and Pankaj Lal[3,4]

[1] Computational Linguistics Program, Montclair State Univ. (MSU)
[2] School of Computing, Montclair State University
[3] Dept. of Earth and Environmental Studies, Montclair State University
4 Clean Energy & Sustainability Analytics Center, Montclair State Univ.
Montclair, New Jersey (NJ) 07043, USA
(bittencourti1 | vardea | lalp)@montclair.edu

**Abstract.** Renewable energy sources are vital to help mitigate the effects of climate change, and reducing the carbon dioxide emissions of fossil fuels, e.g. the state of New Jersey has a goal of producing 100% clean energy by 2050. However, the plans for offshore wind energy by the shore of the state still brings much controversy between residents due to the wind farms' impact on wildlife, coastline, and the people's view from the beaches. In this context, we perform sentiment analysis on social media data to investigate people's opinions and concerns regarding offshore wind energy. We adapt 3 machine learning models, i.e. TextBlob, VADER and SentiWordNet for sentiment analysis because different functions are provided by each model, all of which are useful in our work. Techniques in NLP (natural language processing) are harnessed to gather meaning from the textual data in social media. Data visualization tools are suitably deployed to display the overall results. Despite the controversy surrounding this topic, our findings indicate some positive reception, suggesting potential support for modern-day renewable energy goals. However, there are neutral and negative comments as well, thus potentially helping to find areas for further improvement. The results of this work can be thus useful in a variety of decision-making contexts by governmental organizations and companies, hence aiding and enhancing offshore wind energy policy development. Hence, this work is much in line with citizen science and smart governance via involvement of mass opinion in decision support. In our paper, we highlight the role of sentiment analysis from social media in this aspect.

## 1 Introduction

Renewable energy creates shared value, circular economy approaches and commitment to the United Nations Sustainable Development Goals (UN SDG). Producing more renewable energy and abandoning conventional sources, such as coal, natural gas and petroleum, is a need shared by all countries in the world, especially as global warming grows more intense. Renewable energy has become more popular over the years; and hence many countries, e.g. the United Kingdom, Germany and China have adopted it as a source of clean energy.

Offshore wind energy is a type of renewable energy; it is obtained from the use of the force of the wind on the high seas, where it reaches much higher and more constant speed compared to onshore wind, due to the lack of barriers. According to Environmental America [1], the United States has the potential to achieve the production of more than 7,200 terawatt-hours (TWh) of electricity from offshore wind, which is almost twice the amount of electricity the nation consumed in 2019, and almost 90% of the amount of electricity it would consume in 2050.

Many regions in the United States are coastal states, such as New Jersey, which makes them an ideal location for offshore wind farms. Fig. 1 is a group of pictures taken at ACUA, a wastewater treatment facility in Atlantic City, New Jersey.

During our own site-visit to the facility, we have learned that the plant requires 2.5 megawatts of power each day, and when the speed of the wind is above 12 miles per hour, the electricity they need for a day can be generated by 2 of the 5 turbines they have on site [2]. Considering the potential of these turbines in producing high amounts of clean energy, the state of New Jersey has established a goal of having 100% clean energy by 2050, and 7,500 megawatts coming from offshore wind by 2035 [3].

Although this type of energy is clean and renewable, it raises a lot of concerns regarding wildlife, tourism, fisheries, and coastal communities. For instance, in the state of New Jersey, the Protect Our Coast NJ website is an initiative from the community to stop the building of wind farms, as can be seen in Fig. 2. Accordingly, Fig. 3 is a sample of two posts from a New Jersey Facebook group, where people can post and debate on the topic of offshore wind energy. The wind farm process plan includes “create participation opportunities and resources that address resident concerns in relation to livelihood, landscape, and property / ownership types” [4].

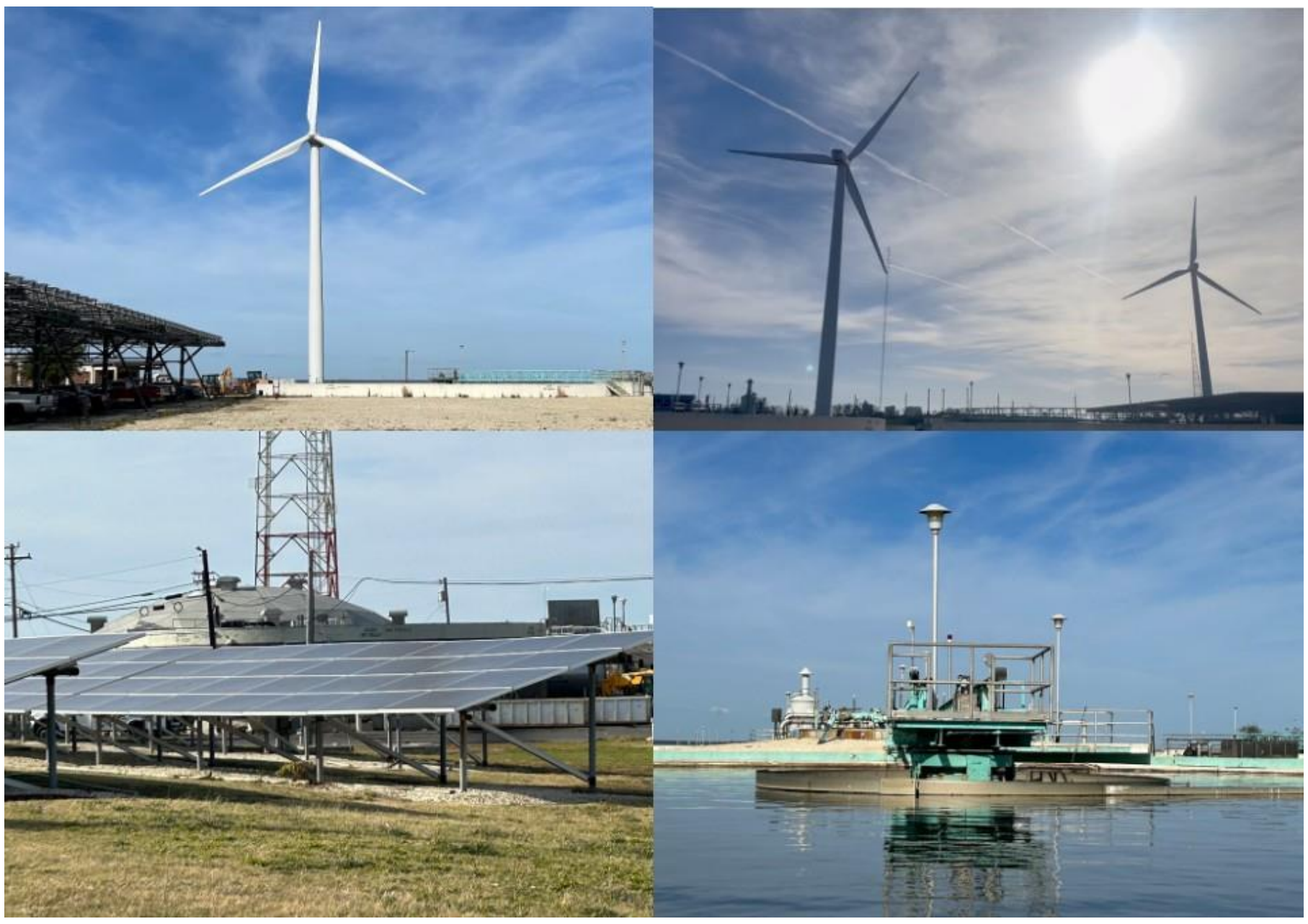

**Fig. 1.** Pictures taken during the Offshore Wind Energy Fellowship meeting at ACUA, wastewater treatment facility in Atlantic City. On the top, working wind turbines. On the bottom left, solar panels. On the bottom right, clarifier tank used for sedimentation.

Hence, given this background and motivation, our study in this paper investigates people’s sentiment towards offshore wind energy through sentiment analysis, offering insights on media opinion to help in fostering the effectiveness of various communication strategies, and also to identify controversies and assess them, thereby supporting the growth of the offshore wind energy initiatives. The main focus of our study in this paper is the state of NJ where we have made site visits, e.g. as seen in Fig. 1 here. Note that the opinion mining conducted here can have broader implications because similar views can be expressed by masses elsewhere in response to renewable energy sources.

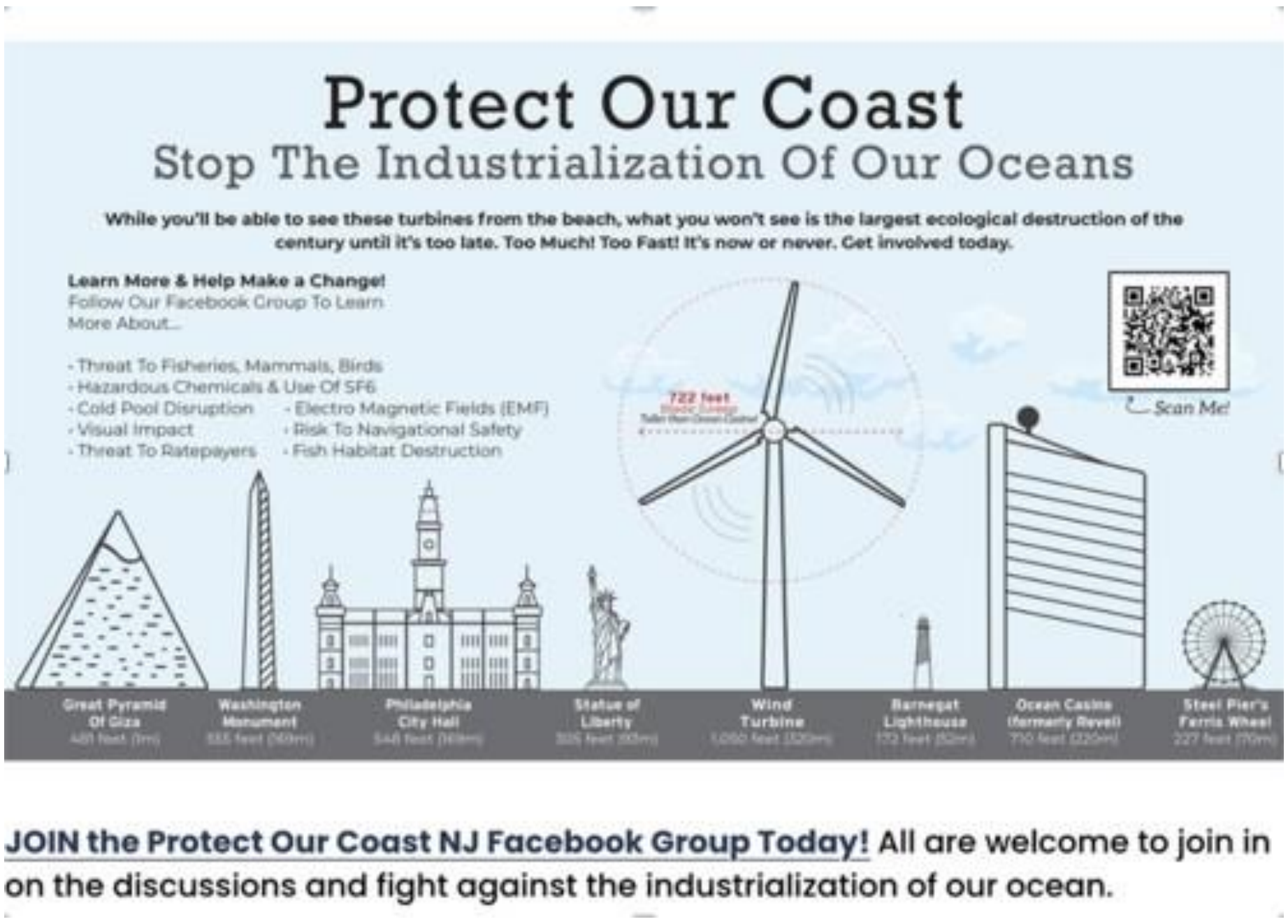

**Fig. 2.** Poster taken from a New Jersey community website w.r.t. renewable energy

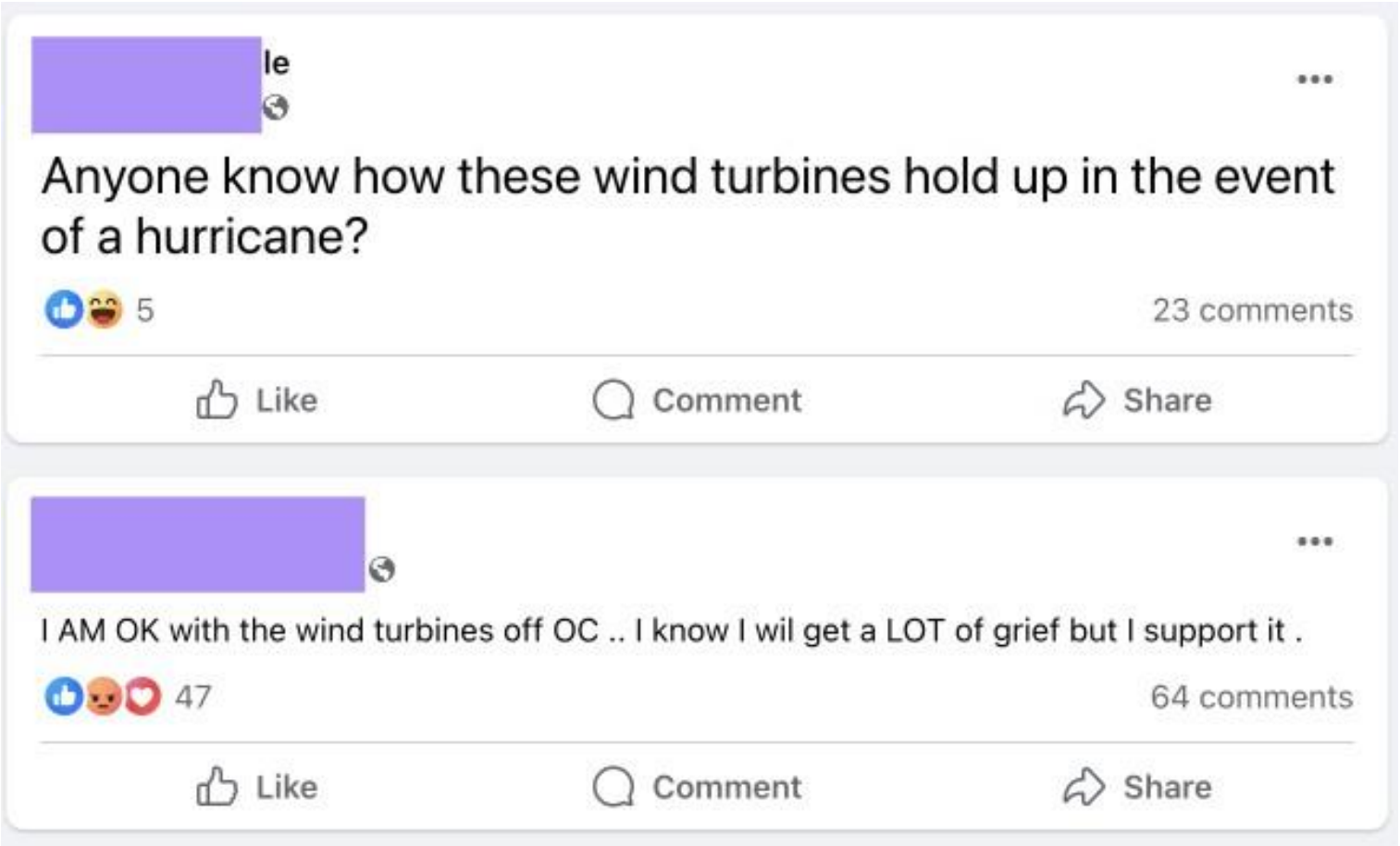

**Fig. 3.** Examples of social media posts from a community group on Facebook.

## 2 Related Work

Sentiment analysis has surfaced as a very popular method of analyzing mass opinion; and consequently, a significant amount of work has been done on urgent and emerging subjects using machine learning models.

In [5], [6] sentiment analysis is used to analyze sentiment towards climate change on Twitter. Sentiment analysis is also performed in [7], where the authors analyze the legitimacy of wind power in Germany using newspaper articles from 2009 to 2018. In [8], sentiment analysis is used for analyzing a large dataset of geo-tagged tweets to investigate the main topics of discussion and sentiment of the population regarding climate change in different countries over time. In [9], the authors perform sentiment analysis and topic modeling on online reviews for policy amendments, showing how air traffic perceptions were impacted by COVID. In [10], opinion mining is used to investigate public reactions towards urban ordinances for smart governance.

Work has also been conducted towards the policies regarding offshore wind energy to investigate major concerns that were already previously known. In [11], the authors provide an overview of the current development of offshore wind power in different countries, and explore issues around its development. The authors in [12] review three areas that can help with reducing energy consumption, IoT, cloud computing and opinion mining, the latter being significant for its contribution when the aim is to understand feelings and demands from energy consumers and stakeholders to help in the creation of better policies. Additionally, in [13], the authors present an overview of the main issues associated with the economics of offshore wind. In [14], the costs and benefits of offshore wind are discussed relative to onshore wind power and conventional electricity production, with a review of cost estimates.

## 3 Data Description

The data used in this project is gathered mainly from Facebook groups. It has become very common for towns to create groups for discussing topics about their community. We aptly target communities from the NJ coast, e.g. groups from Atlantic City, Ocean City, The Wildwoods and Cape May. We only collect data from publicly accessible groups, so that there are no privacy concerns. We use the Apify web scraper platform [15] to collect and create the dataset. Comment samples can be seen in Fig. 3 and 4.

We use standard search engines to find pertinent comments related to the keywords “offshore wind energy”, “wind farms” and “wind turbines”. A total of 6569 comments are collected, constituting the main source of data in our analysis. Note that we focus our analysis on Facebook posts, however our models and algorithms can be used for other social media sources for opinion mining in related contexts.

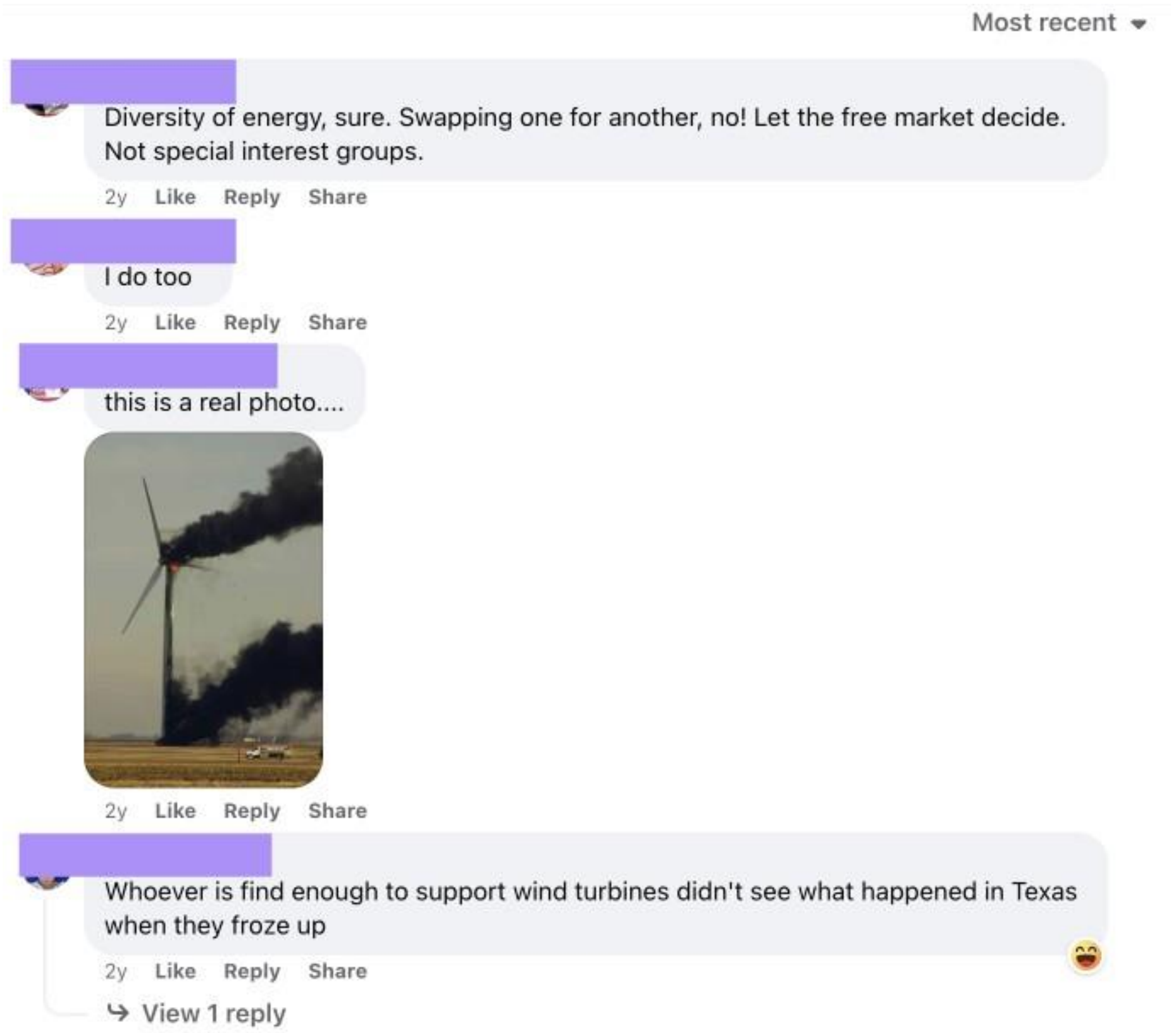

**Fig. 3.** Sample 1 of comments from a Facebook NJ community group.

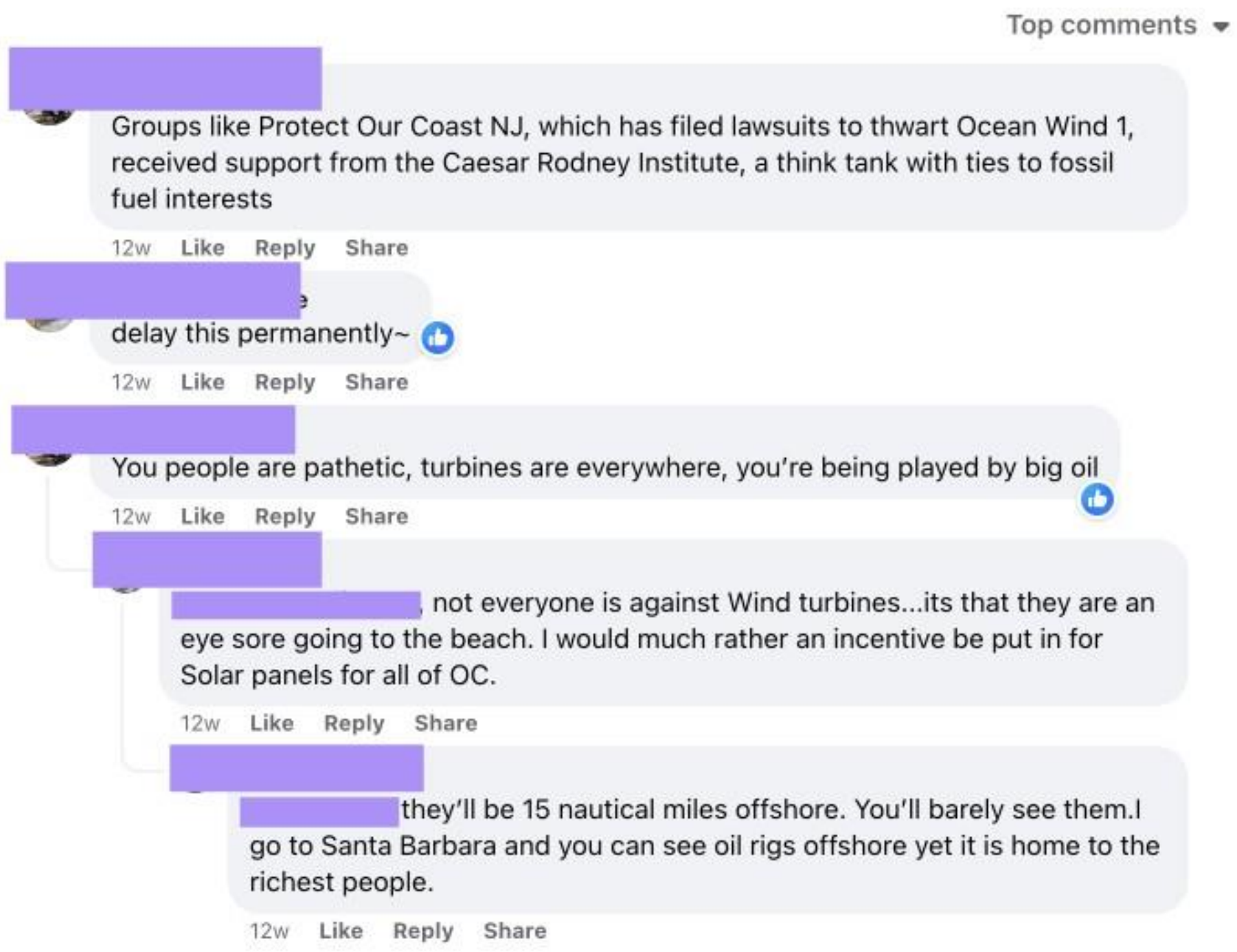

**Fig. 4.** Sample 2 of comments from a Facebook NJ community group.

## 4 Details of Methods

Our overall approach for sentiment analysis is synopsized with an illustration in Fig. 5. After collecting the data, our first step is preprocessing. We use the Python NLTK (natural language toolkit) library [16] to remove punctuation, html links and stop words, and to perform lemmatization and stemming. It is important to use these preprocessing techniques before analyzing text with sentiment analysis models, in order to discard irrelevant text, e.g. pronouns, articles, that do not add useful meaning to the analysis. The pseudocode for preprocessing can be seen in Algorithm 1.

---

ALGORITHM 1: Preprocessing social media data on offshore wind

---

```
INPUT: Facebook comments data-frame δ
FOR comment α in δ DO:
    IF α is null:
        Remove α from δ
    END IF
    Make α lowercase
    Delete punctuation in α
    Delete URLs from α
    Delete # from α
    DO stop-word elimination on α
    DO Lemmatization on α
    IF α is < 3:
        Remove α from δ
    END IF
END FOR
OUTPUT: preprocessed text data as cleaned text β
```

---

Once the data is preprocessed, we commence the core of the analysis. We deploy the paradigm of sentiment analysis, i.e. a series of methods, techniques and tools used to detect and extract subjective information, such as opinion and attitudes, from language. This paradigm suits our goals of identifying subjective information on offshore wind energy. As is widely known, the recent advances in deep learning and the ability of algorithms to analyze text, have made sentiment analysis improve significantly. Thus, its practice has increased tremendously to gauge mass opinions.

In order to perform sentiment analysis on our data, we use 3 models: TextBlob [17], VADER (Valence Aware Dictionary) [18] and SentiWordNet [19], due to their popularity and overall good performance on sentiment classification in the literature, and due to various functionalities provided by these models, all of which are useful in the context of our work. All these models return polarity scores for each comment, which allow us to tag each comment as positive if the score is a positive number, negative if the score is a negative number, and neutral if the score is equal to zero. We make the decision to use these 3 different models because each model outputs polarity scores differently as explained next. The pseudocode for obtaining polarity analysis in our work is displayed in Algorithm 2.

---

ALGORITHM 2: Sentiment analysis on offshore wind energy posts

---

```
INPUT: Preprocessed Facebook comments data-frame δ
FOR comment α in δ DO:
    IF polarity score φ > 0 DO:
        Add positive label
    IF φ = 0 DO:
        Add neutral label
    ELSE DO:
        Add negative label
    END IF
END FOR
OUTPUT: polarity scores φ for each comment α
```

---

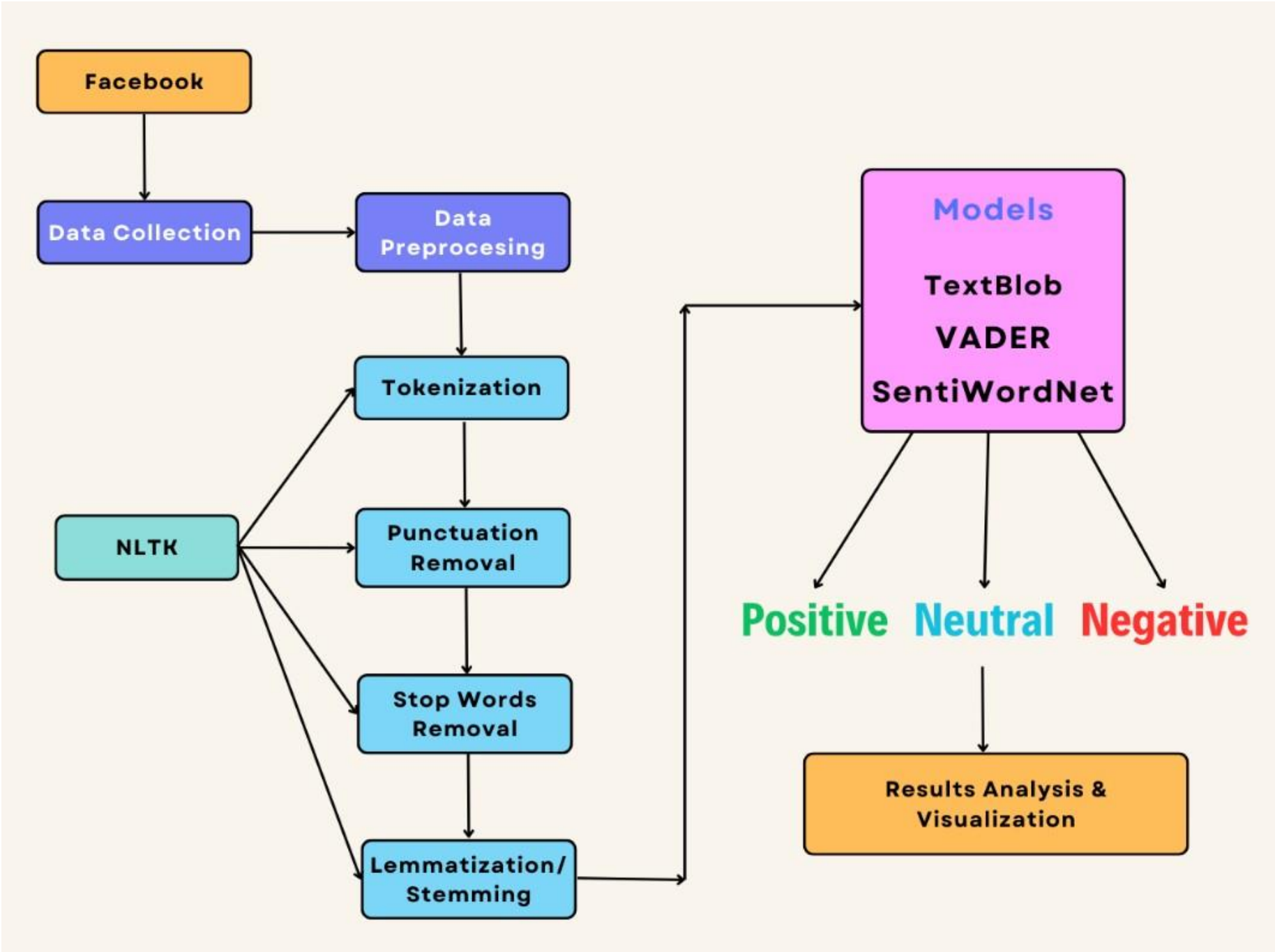

**Fig. 5.** Approach for Sentiment Analysis

TextBlob gives polarity scores by aggregating individual words' sentiment scores. Besides returning polarity scores, TextBlob also returns subjectivity scores. The subjectivity scores lie between 0 and 1 and show if the text expresses subjective or objective content. The higher the subjectivity means that the text has more opinions, emotions or personal judgements (rather than factual or more neutral content which yields a lower subjectivity score). For example, if a post contains the text, "I do not like offshore wind energy, it's boring!", it is really subjective and opinionated, and would thus get a very high subjectivity score. On the other hand, if it has the text, "Offshore wind energy costs us 10% more than our current usage which we cannot afford due to our profits being 20% lower this year", the information is more objective and factual, hence its subjectivity score would be lower. Although we are looking into people's opinions towards offshore wind energy, TextBlob's subjectivity feature helps us understand the type of information that is being disseminated in the comments, thereby providing a clearer insight into the authenticity of the comments. Besides opinionated comments, factual comments can show us that people are actually discussing facts about the topic. This is our main reason for deploying TextBlob to get a good idea of the subjective versus objective content in the posts.

VADER (Valence Aware Dictionary and Sentiment Reasoner) returns normalized polarity scores. VADER's compound scores are calculated using the sum of positive, negative, and neutral scores, which are normalized between -1 (most negative) and 1 (most positive). For instance, if a user comments that "Renewable energy sources maybe a bit expensive but are much healthier", it can get a highly positive score for "much healthier" but a slightly negative score for "bit expensive" yielding a compound score that tilts more towards the positive side. Hence, we harness VADER for compound scores to enable a more cumulative analysis of the posts.

SentiWordNet thrives on the well-known lexical source WordNet and considers the given context using "synsets". These synsets are sets of synonyms that are grouped together by their semantic equivalence, which are useful for analyzing the semantic context of the textual data. For example, the word "estimable" would get a neutral score when referring to an item because it implies an item that can be estimated (e.g. its cost); while the same word "estimable" would receive a positive score when referring to a person because it typically refers to a person worthy of esteem or respect. This justifies our choice of SentiWordNet to facilitate adequate analysis with reference to context in the posts considered.

## 5 Experimental Results

The sentiment analysis performed in this work shows that the overall sentiment on offshore wind energy seems fairly positive on the whole across all the 3 models, TextBlob, VADER, and SentiWordNet; however, there are many users posting negative and neutral comments as well. Figs. 6, 7 and 8 present the sentiment distribution as obtained in TextBlob, VADER and SentiWordNet respectively. All these figures include bar plots and pie plots for at-a-glance visualization of the results derived from each model. Delving deeper into the posts pertaining to the negative and neutral comments can pave the way to identify loopholes in existing policies based on their reception by the masses, thus presenting potential areas for improvement. Likewise, the positive posts can offer the scope for continuation of the respective policies and further encouragement along similar lines.

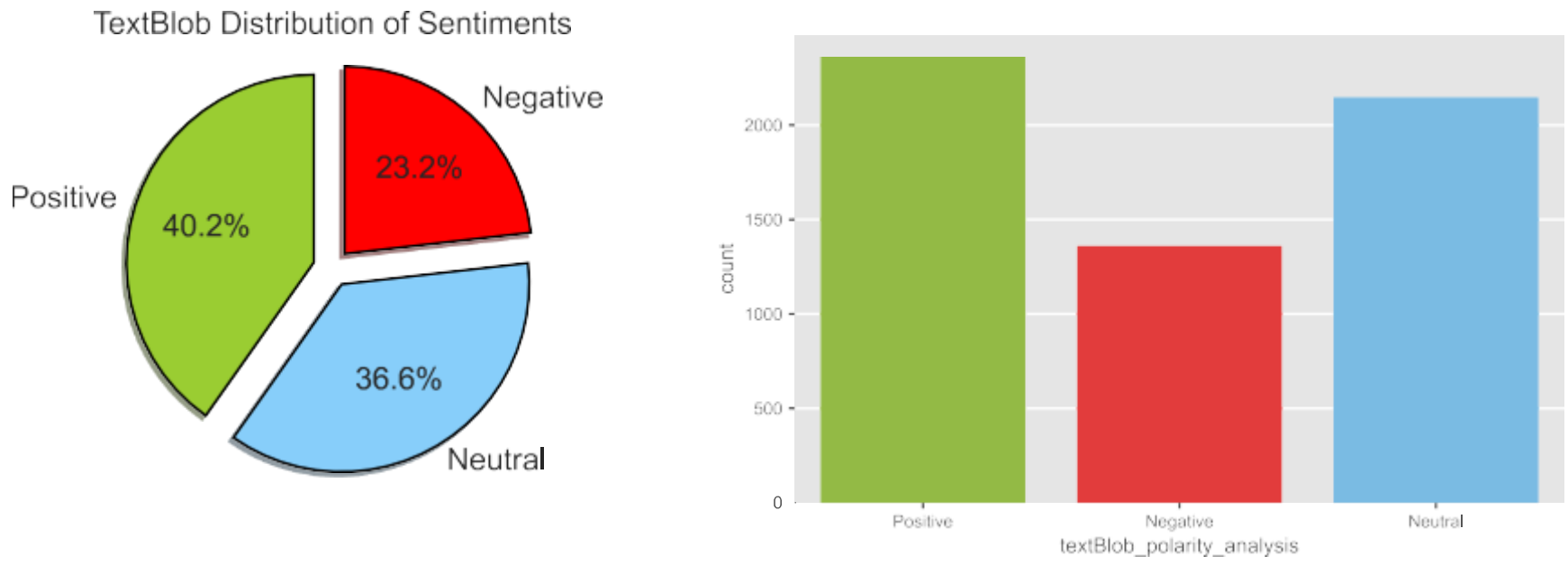

**Fig. 6.** TextBlob plots for distribution of sentiments visualization.

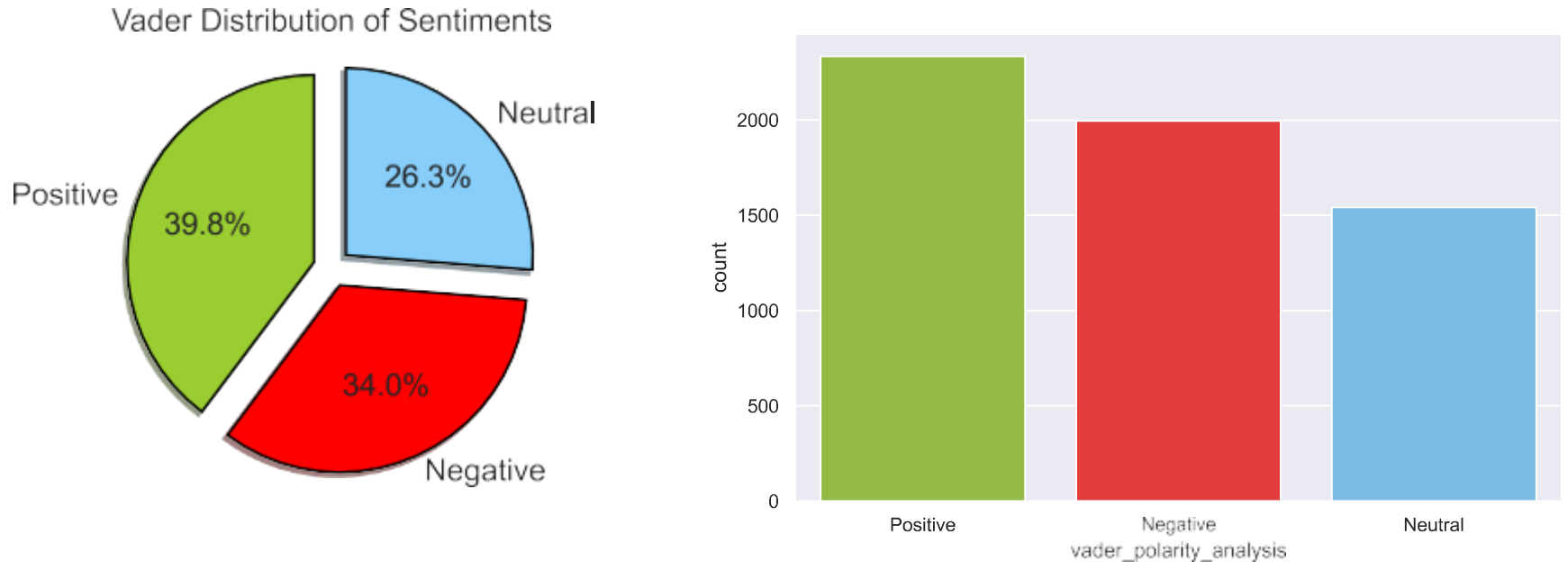

**Fig. 7.** VADER plots for distribution of sentiments visualization.

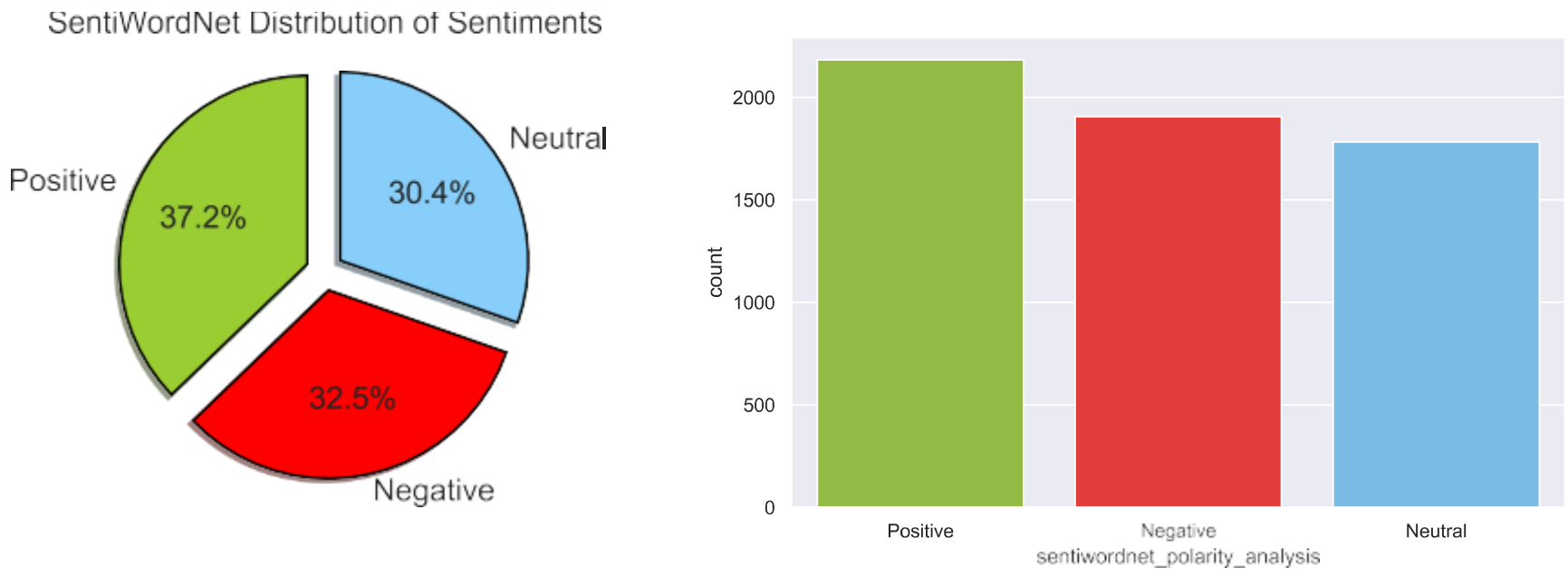

**Fig. 8.** SentiWordNet plots for distribution of sentiments visualization.

As discussed earlier, we are able to get subjectivity scores from TextBlob. In Fig. 9, we show the subjectivity scores within the dataset in a bar plot. It is possible to see that the content of the dataset is mostly factual, which indicates that besides expressing their opinion, people are discussing real facts within their comments.

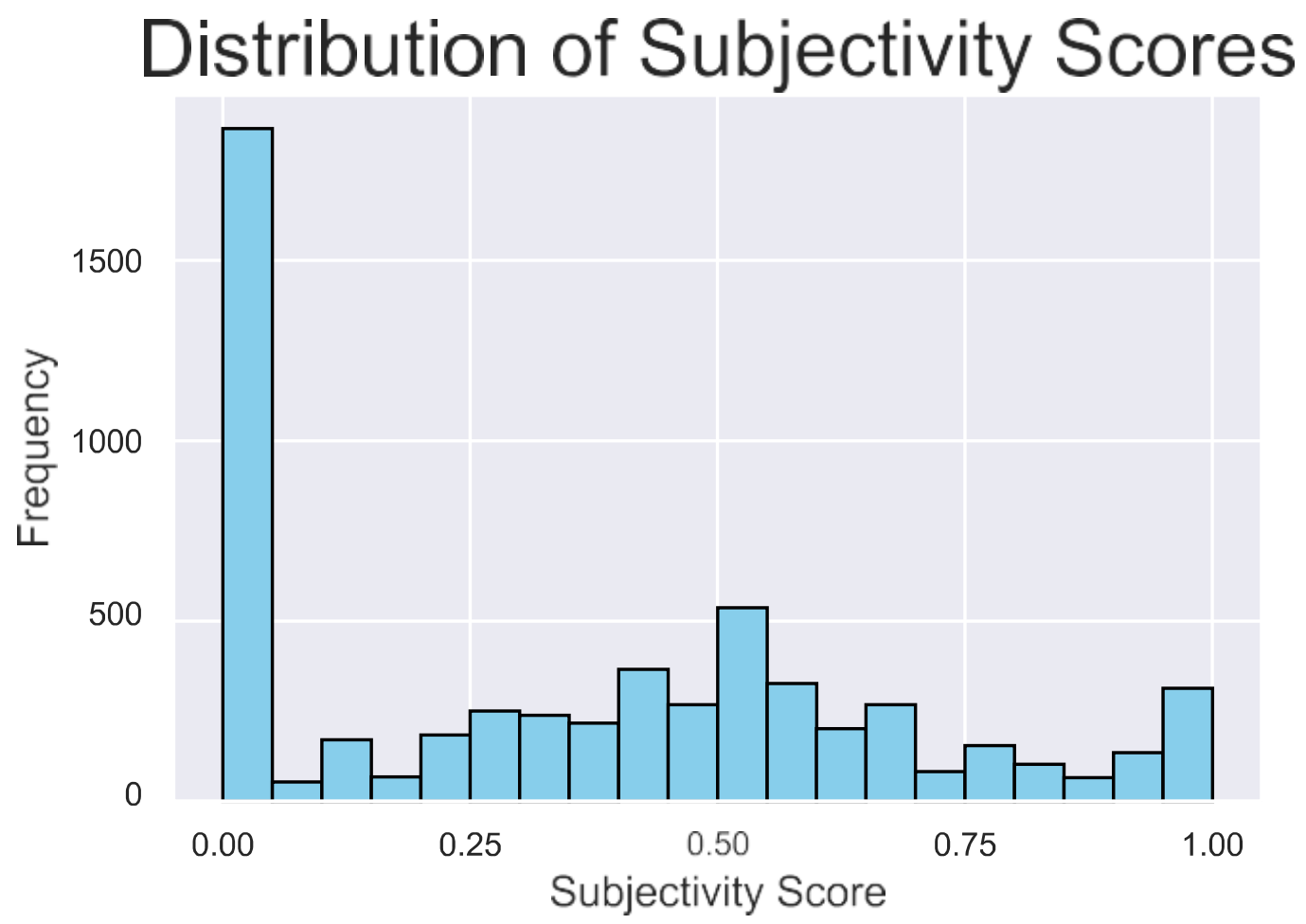

**Fig. 9.** TextBlob plots for distribution of subjectivity visualization.

Moreover, it is possible to visualize the most frequent negative and positive words present in the comments. We extract the most frequent negative words in the negative comments and the most frequent positive words in the positive comments from each model. These are plotted in Figs. 10 and 11 for TextBlob, Figs. 12 and 13 for VADER, and Figs. 14 and 15 for SentiWordNet, respectively. These serve to provide good visual depictions of the main terms of interest in the social media posts.

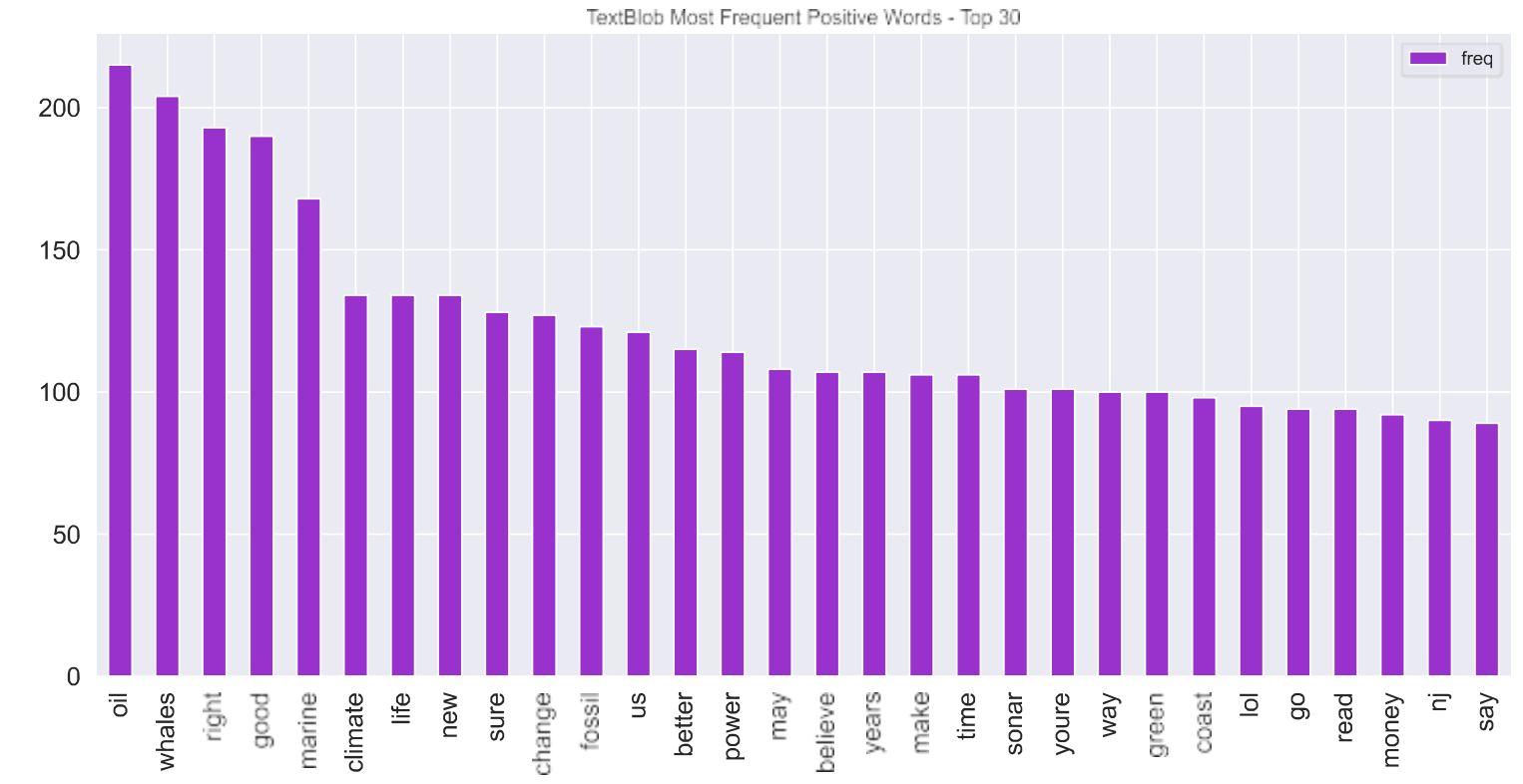

**Fig. 10.** TextBlob's top 30 positive words.

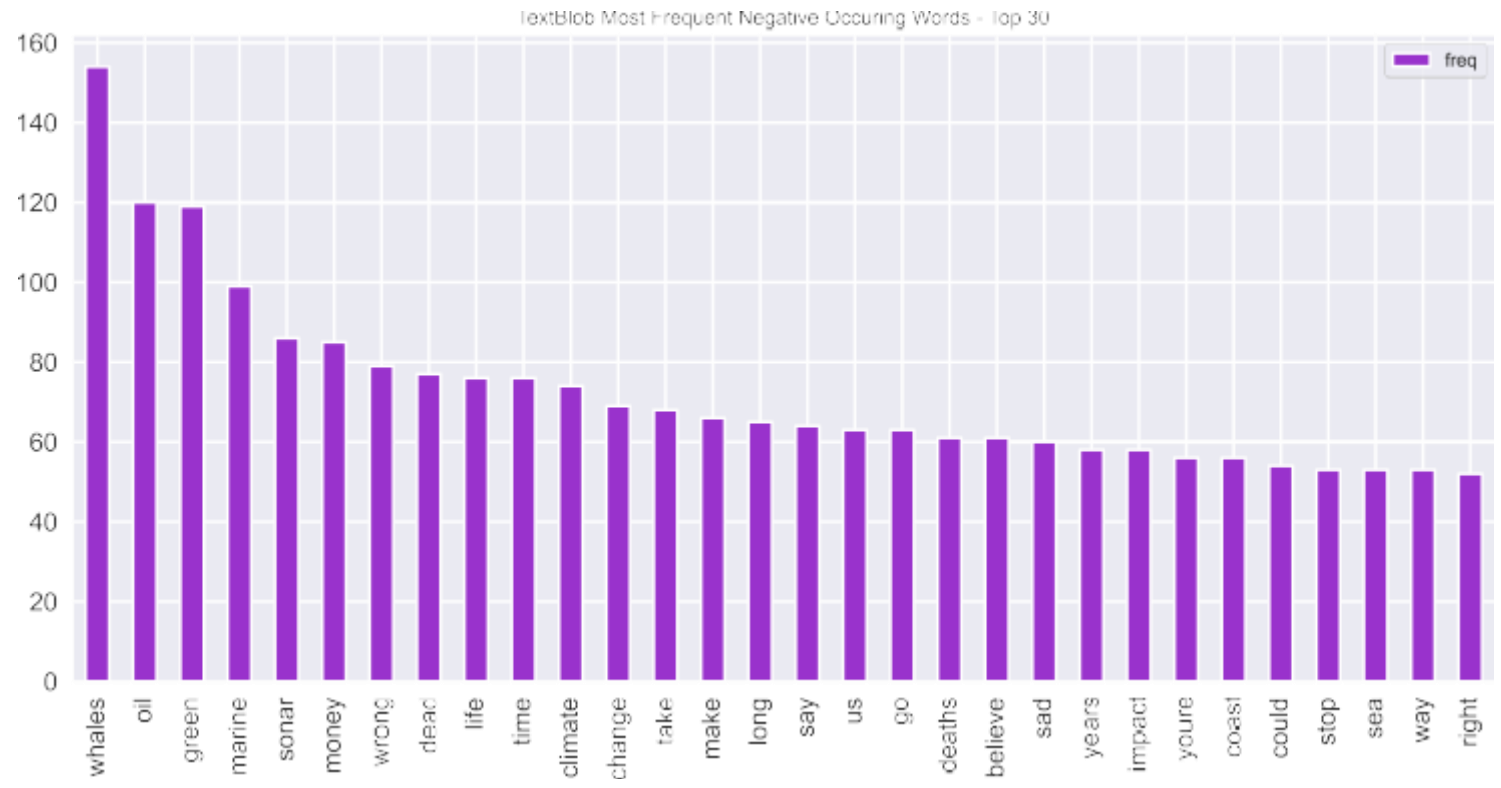

**Fig. 11.** TextBlob's top 30 negative words.

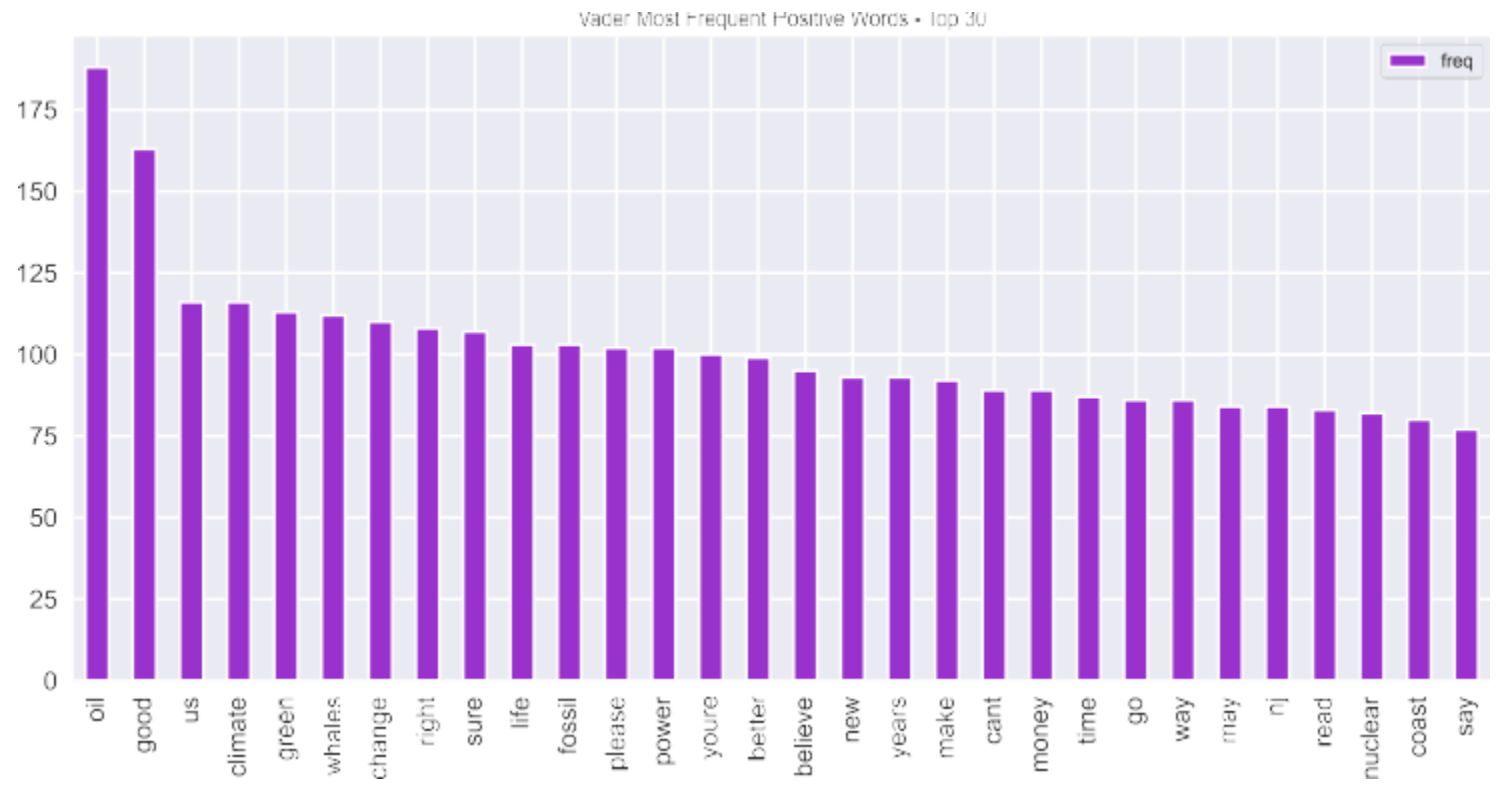

**Fig. 12.** VADER's top 30 positive words.

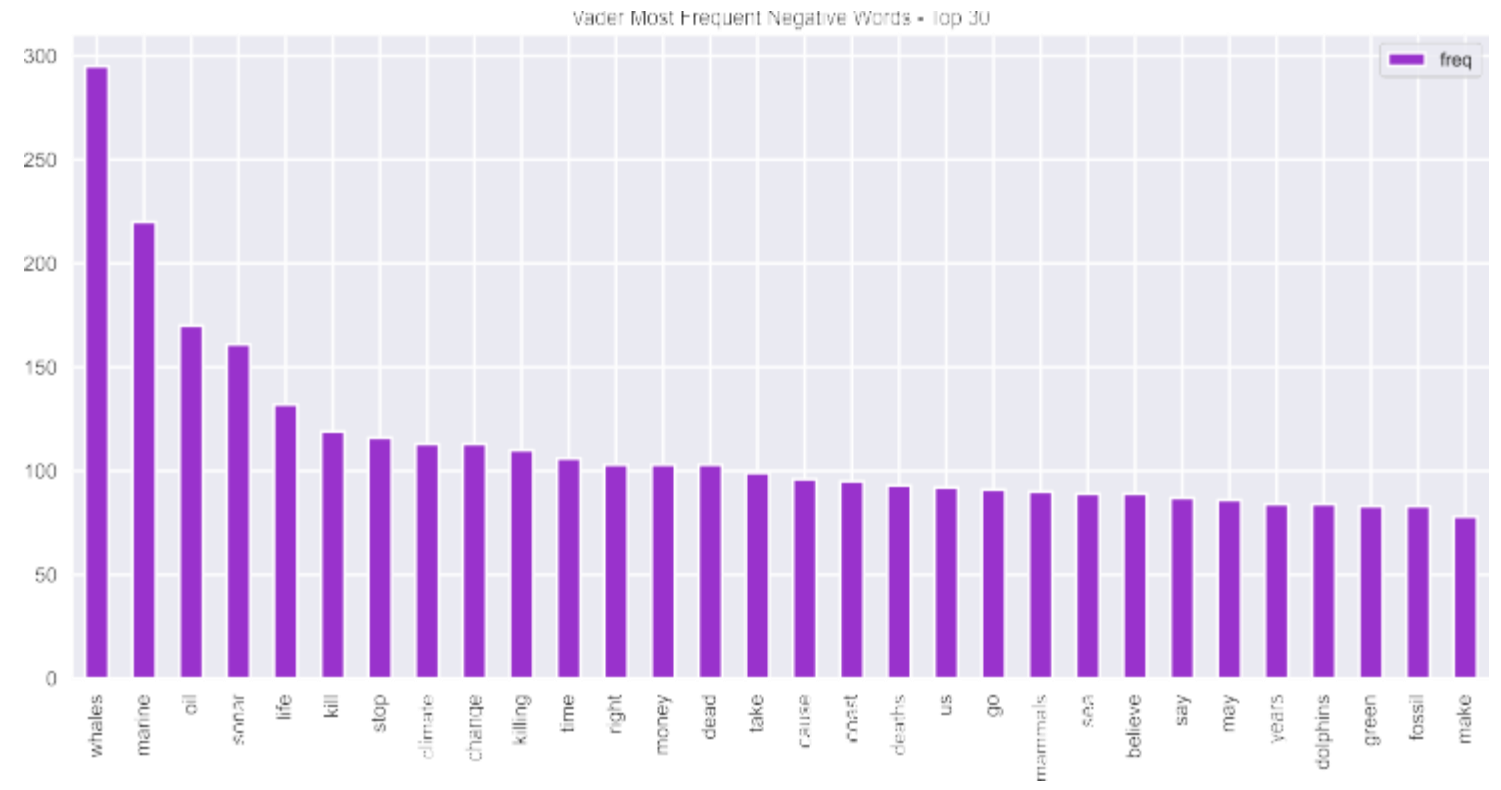

**Fig. 13.** VADER's top 30 negative words.

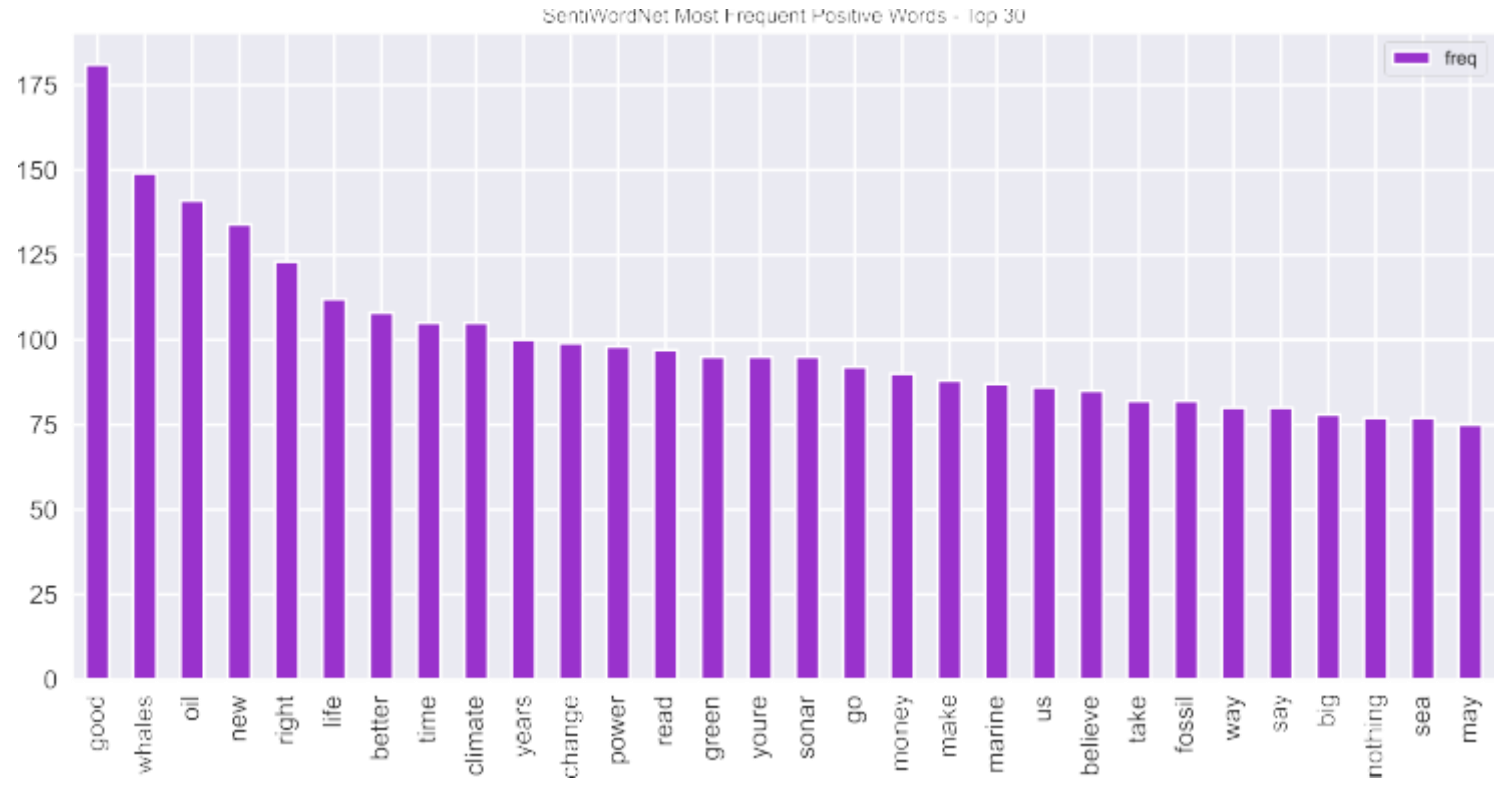

**Fig. 14.** SentiWordNet's top 30 positive words.

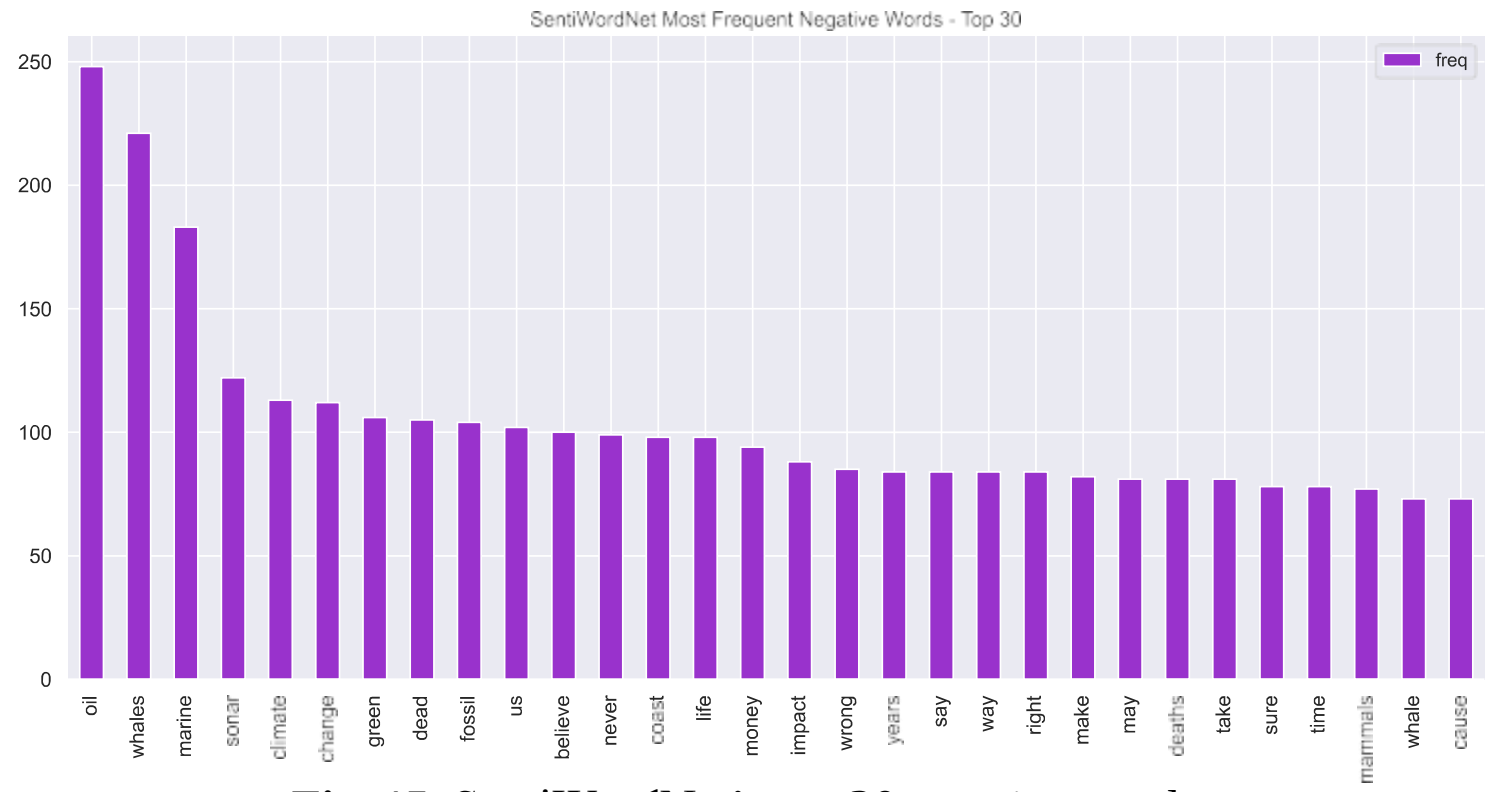

**Fig. 15.** SentiWordNet's top 30 negative words.

Frequent words reveal major topics of interest in the data. Some highly significant topics of discussion in this analysis can be seen in the positive and negative WordClouds in Figs. 16 through 21, created using the most frequent positive and negative words from each model. The bigger the word appears in the image, the more frequent it is. A few significant words are marine mammal, Osterd, whale, fossil fuel, propaganda, climate change, wildlife, money, democrats, sonar mapping, death etc.

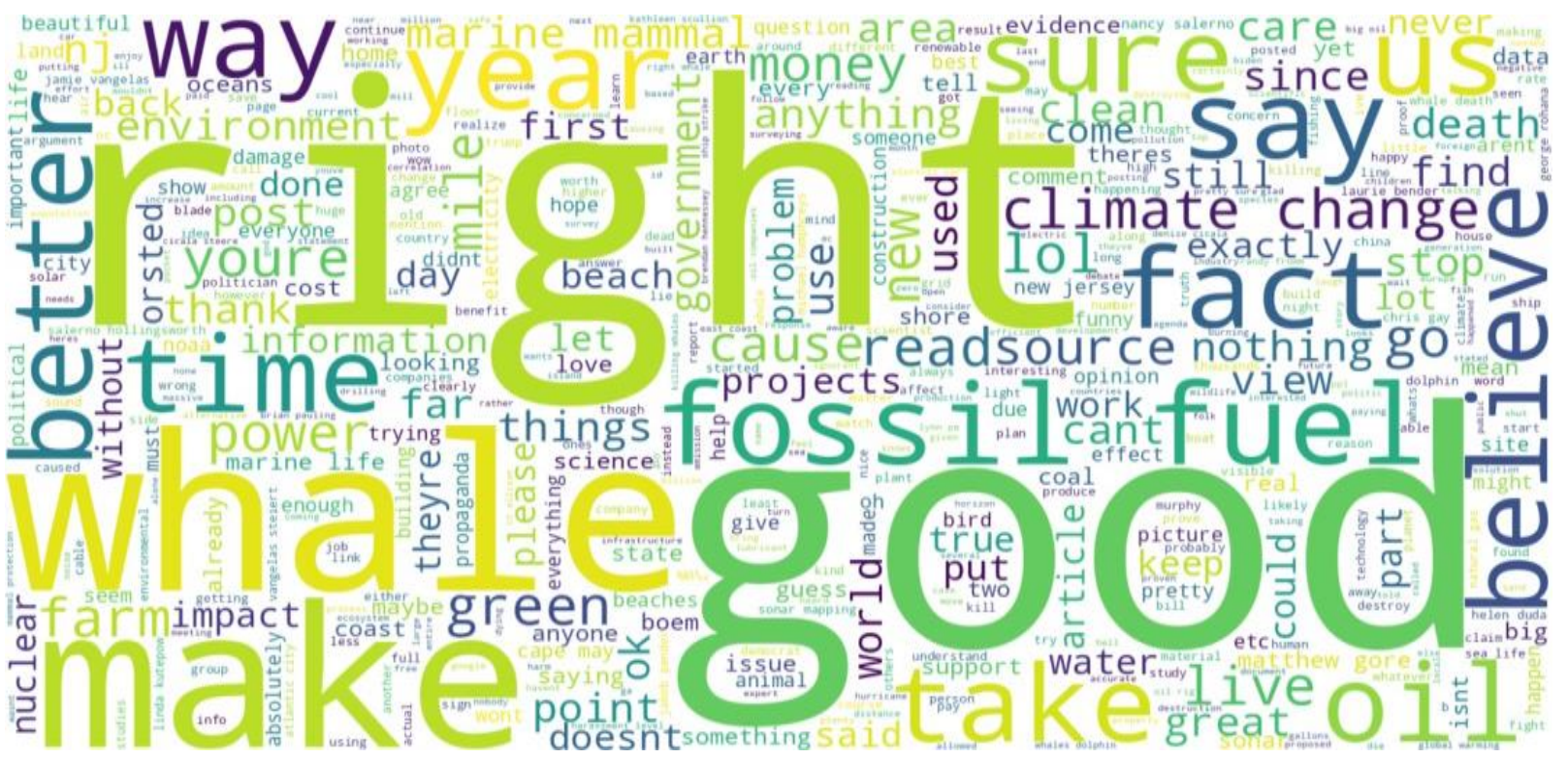

**Fig. 16.** Positive WordCloud from TextBlob.

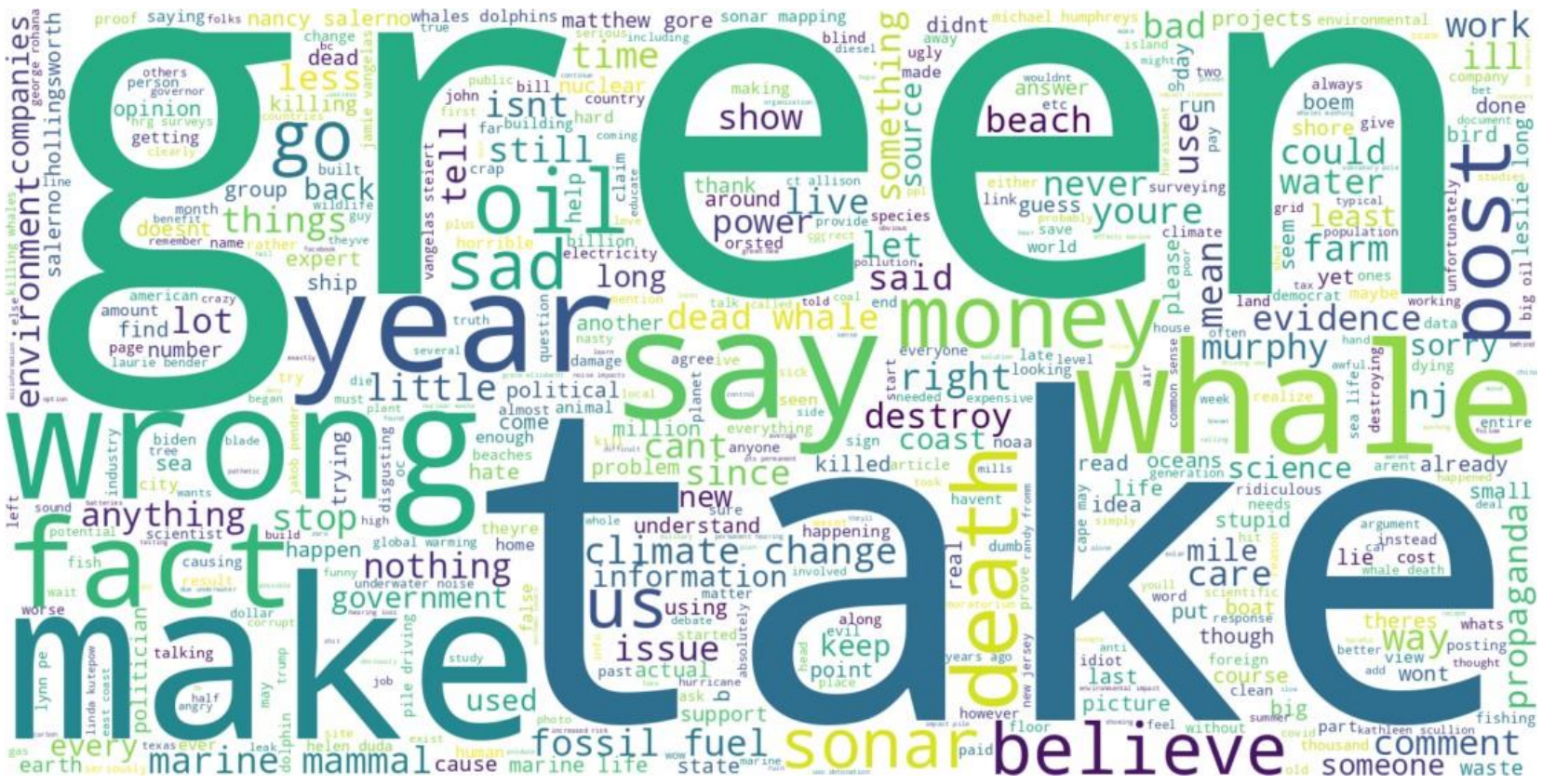

**Fig. 17.** Negative WordCloud from TextBlob.

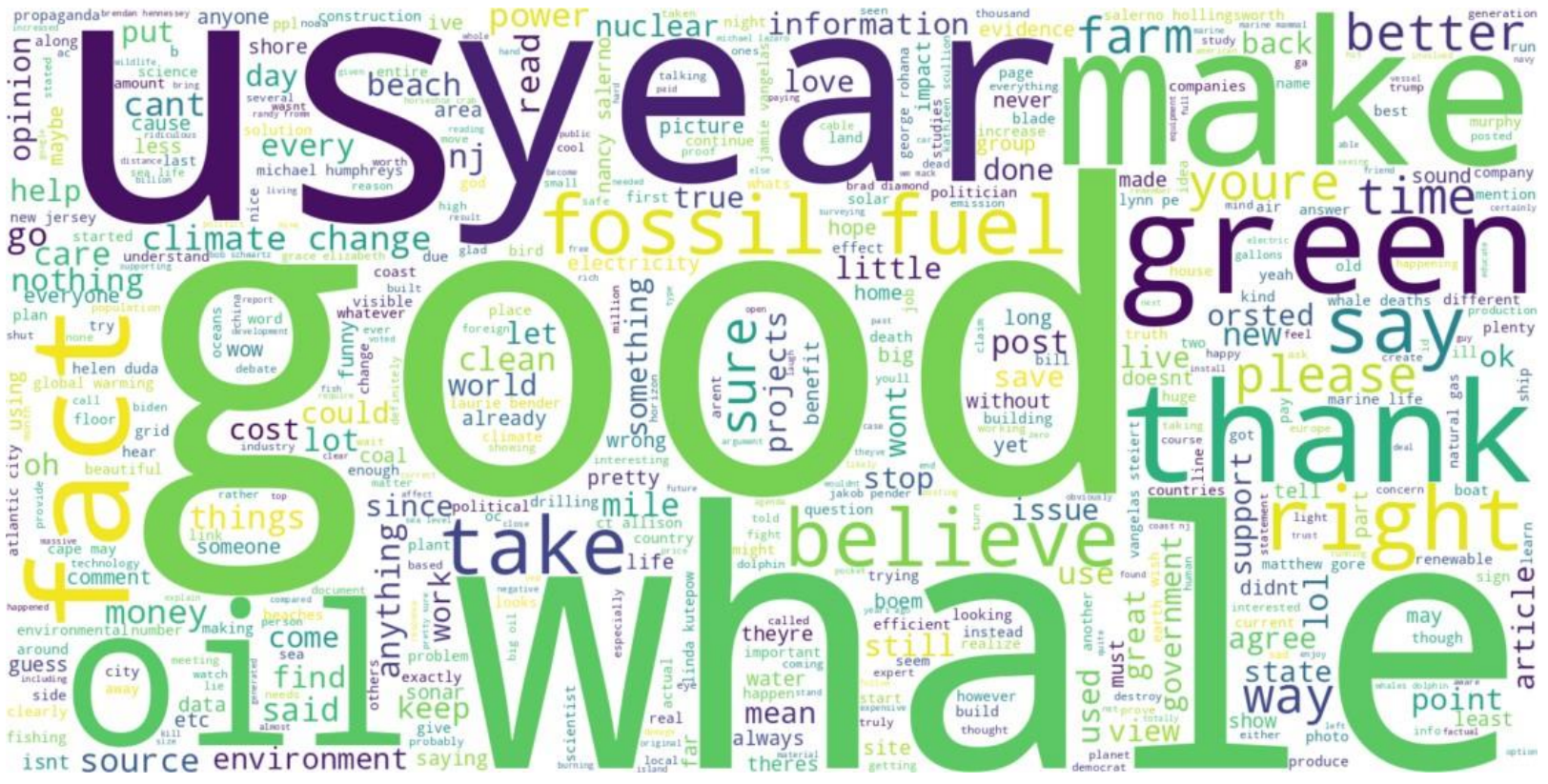

**Fig. 18.** Positive WordCloud from VADER.

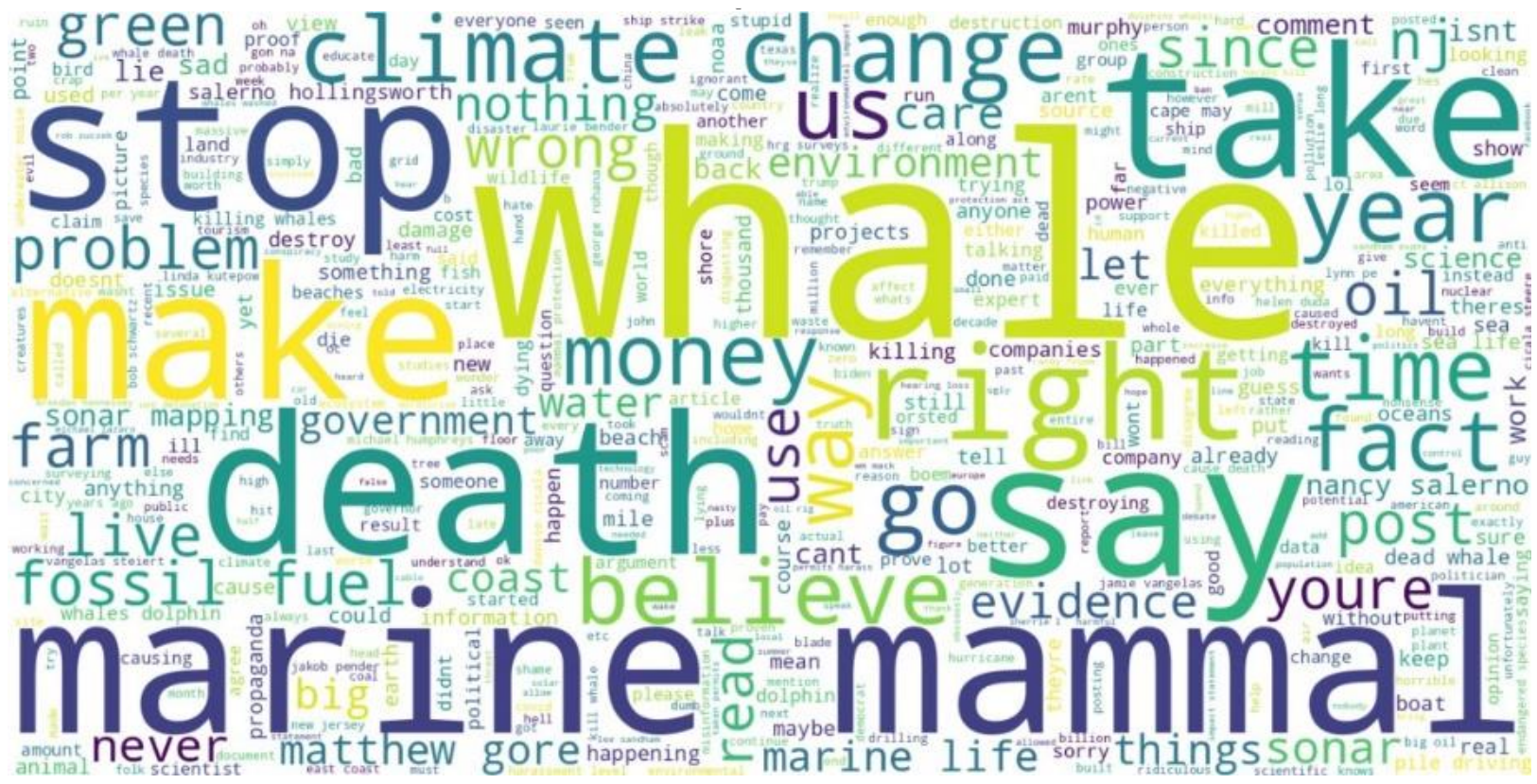

**Fig. 19.** Negative WordCloud from VADER.

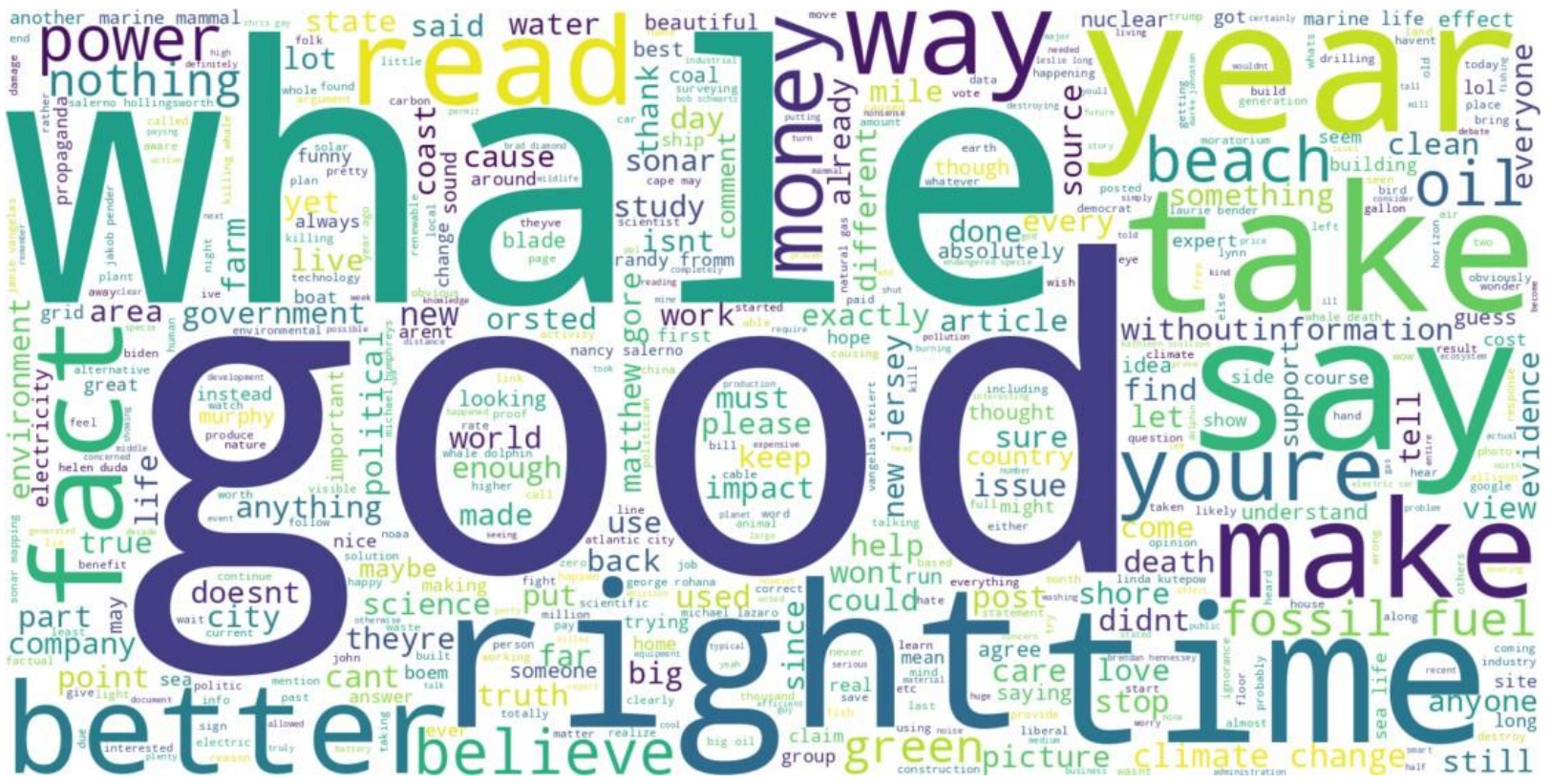

**Fig. 20.** Positive WordCloud from SentiWordNet.

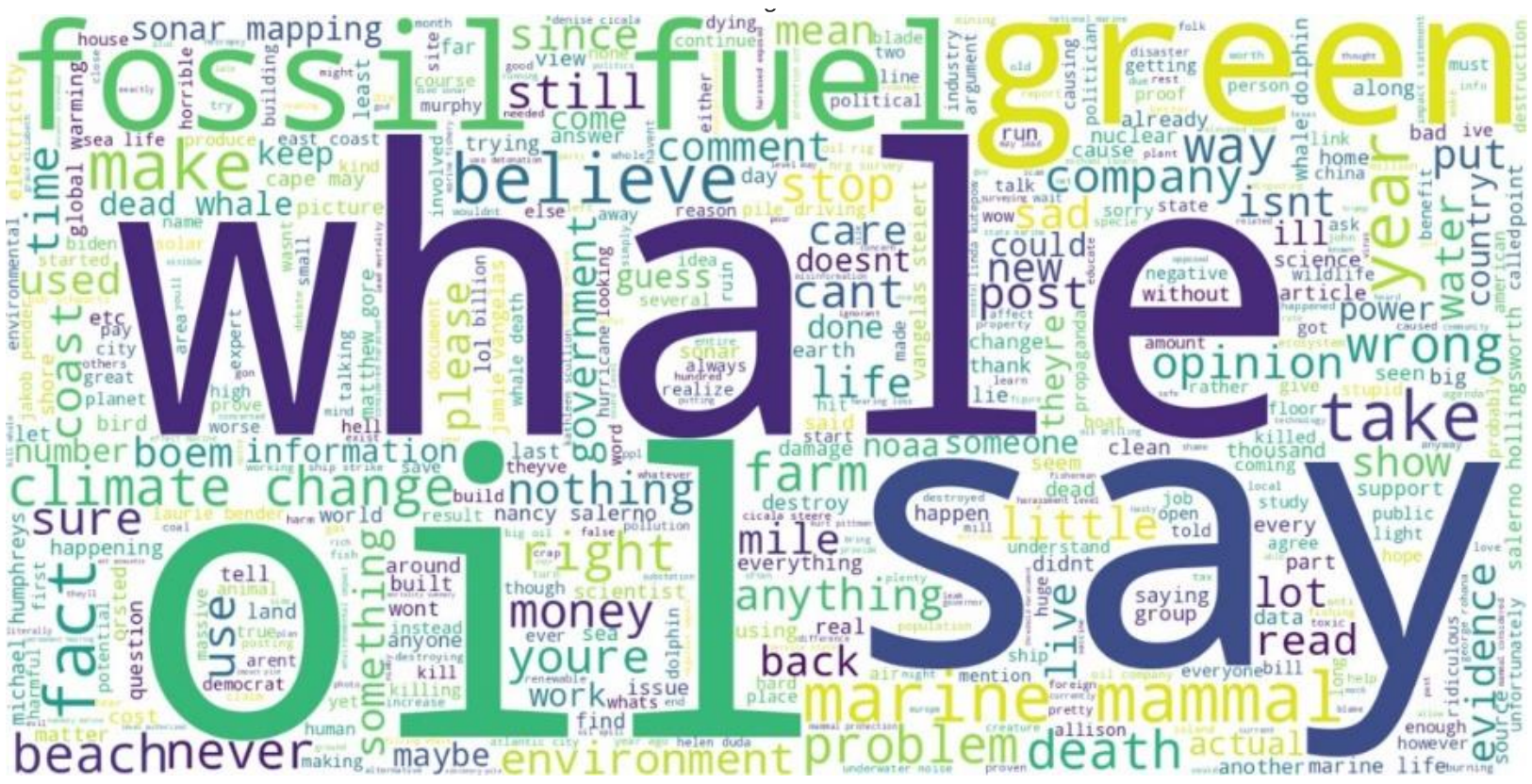

**Fig. 21.** Negative WordCloud from SentiWordNet.

## 6 Conclusions and Future Work

This work harnesses methods in machine learning and natural language processing to address issues in environmental management, more specifically, clean energy. In particular, offshore wind energy is targeted here, in order to gauge its reception by the masses. Sentiment analysis is performed on social media data, more specifically Facebook posts geo-located in NJ, to investigate people's opinions. The need for renewable sources of energy is high due to the severe effects of climate change on earth. This study provides support to the commitment of producing 100% clean energy by 2050 that is being targeted across many regions in the world today, e.g. in the state of New Jersey here. Although the overall opinion of the population in this dataset is somewhat positive, the neutral and negative counts are not far behind due to concerns over the impacts of offshore wind energy on wildlife and beaches, the state's budget money, etc. Subjectivity scores show that people have a considerable extent of concrete and factual discussions in their social media posts, which is an interesting finding.

The results presented here can be used for future decisions by government leaders and interested companies on the topic, i.e. wind farms' location, projects budget, environmental impacts etc. Tailoring policies to public opinion and identifying key concerns are an important part of the process of investing in the build of wind farms by the coast of New Jersey.

Likewise, similar analyses can be performed on social media posts in other regions to gain deeper insights on the reception of the policies by the common masses. This is in line with paradigms such as citizen science and smart governance that entail active involvement of the masses with more transparency and openness, providing the foundation for more adequate decision-making. Our work in this paper thus makes a modest contribution to these paradigms.

Future work emerging from our research includes investigating other sources of text such as online news, corporate and governmental websites, user blogs, and research articles to discover knowledge from text. This can be achieved via machine learning techniques in conjunction with natural language processing. It can include topic modeling using LDA (Latent Dirichlet Allocation), for instance as in [20] where LDA is used to explore climate change, energy and food security trends in newspapers and public documents. It can also entail analyzing more information with Large Language Models (LLMs), such as in [21], where the authors use LLMs for topic modeling and perform quite well. Furthermore, it is possible to map text to structured data to enable the usage of data mining techniques such as association rules for enhanced knowledge discovery. More research on these lines can be conducted in the broad realm of computational linguistics such that it can be useful in studies on the environment.

In sum, our paper provides the ground for more work on machine learning and NLP techniques to tackle environmental issues, e.g. in clean energy and related areas. As we move forward towards a greener future, there has been an ample amount of research done towards analysis of environmental topics, deployment of environment-friendly applications, scientific data mining in general, and machine intelligence on the whole [22, 23, 24, 25, 26, 27, 28, 29, 30, 31] much of which focuses on work done in our research group. Hence, our work in this paper is orthogonal to such research. Overall, our study in this paper stands on a notable bridge between computational linguistics and environmental management, helping in the achievement of more sustainable practices for a better and greener world.

## 7 Acknowledgments

All the authors acknowledge the New Jersey Wind Institute Fellowship Program by NJEDA, a statewide initiative in NJ. We also acknowledge the Clean Energy and Sustainability Analytics Center (CESAC) at MSU, NJ. In addition. Dr. Aparna Varde acknowledges NSF MRI grants 2018575 and 2117308 as well.

## 8 References

1. ENVIRONMENT AMERICA, Offshore Wind for America, https://environmentamerica.org/center/resources/offshore-wind-for-america-3/#:~:text=The%20United%20States%20has%20the,2050%20if%20we%20electrified%20our, last accessed 2023/12/01.
2. ACUA, Jersey-Atlantic Wind Farm, https://www.acua.com/Projects/Jersey-Atlantic-Wind-Farm.aspx, last accessed 2024/02/01.
3. Department of Environmental Protection, OffshoreWind, https://dep.nj.gov/offshorewind/, last accessed 2023/12/01.
4. Elmallah, S., Rand, J.: After the leases are signed, it's a done deal: Exploring procedural injustices for

utility-scale wind energy planning in the United States. Energy Research & Social Science, 89, 102549 (2022).

5. Rosenberg, E., Tarazona, C., Mallor, F., Eivazi, H., Pastor-Escuredo, D., Fuso-Nerini, F., Vinuesa, R.: Sentiment analysis on Twitter data towards climate action. Results in Engineering, vol. 19, pp. 101-287, (2023).
6. McNamee, B., Varde, A., Razniewski, S.: Correlating facts and social media trends on Environmental quantities leveraging commonsense reasoning and human sentiments. In proceedings of The 2nd Workshop on Sentiment Analysis and Linguistic Linked Data, pp. 25-30 (2022).
7. Dehler-Holland, J., Okoh, M., Keles, D.: Assessing technology legitimacy with topic models and sentiment analysis – The case of wind power in Germany. Technological Forecasting and Social Change, vol. 175, pp. 121-354 (2022).
8. Dahal, B., Kumar, S.A.P., Li, Z. Topic modeling and sentiment analysis of global climate change tweets. Social Network Analysis and Mining 9, 24 (2019).
9. Field, A., Varde, A., Lal, P.: Sentiment Analysis and Topic Modeling for Public Perceptions of Air Travel: COVID Issues and Policy Amendments. Language Resources and Evaluation Conference, pp. 2-8, (LREC 2022).
10. Puri, M., Varde, A., De Melo, G.: Smart governance through opinion mining of public reactions on ordinances. IEEE 30th International Conference on Tools with Artificial Intelligence, pp. 838-845 (ICTAI 2018).
11. Sun, X., Huang, D., Wu, G.: The current state of offshore wind energy technology development. Energy, vol. 41, pp. 298-312 (2012).
12. Shrestha, S., Varde, A.: Roles of the Web in Commercial Energy Efficiency: IoT, Cloud Computing, and Opinion Mining. Association for Computing Machinery Special Interest Group on Hypertext, Hypermedia and Web, Article 5, pp. 1-16, (ACM SIGWEB Autumn 2023).
13. Green, R., Vasilakos, N.: The economics of offshore wind. Energy Policy, vol. 39, pp. 496-502 (2011).
14. Snyder, B., Kaiser, M.: Ecological and economic cost-benefit analysis of offshore wind energy. Renewable Energy, vol. 34, pp. 1567-1578 (2009).
15. Web Scraper, Apify, apify.com, last accessed 2023/12/15.
16. Bird, S., Klein, E., Loper, E.: Natural language processing with Python: analyzing text with the natural language toolkit. O'Reilly Media, Inc. (2009).
17. Textblob, textblob.readthedocs.io/en/dev/index.html, last accessed 2023/12/01
18. Hutto, C.J., Gilbert, E.E. VADER: A Parsimonious Rule-based Model for Sentiment Analysis of Social Media Text. Eighth International Conference on Weblogs and Social Media (2014).
19. Baccianella, S., Esuli, A., Sebastiani, F.: SentiWordNet 3.0: An Enhanced Lexical Resource for Sentiment Analysis and Opinion Mining. N. Calzolari, K. Choukri, B. Maegaard, J. Mariani, J. Odijk, S. Piperidis, M. Rosner & D. Tapias (eds.), Language Resources and Evaluation Conference (LREC 2010).
20. Benites-Lazaro, L.L, Giatti, A., Giarolla, A.: Topic modeling method for analyzing actor discourses on climate change, energy and food security. Energy Research & Social Science, 45:318-330 (2018).
21. Stammbach, D., Zouhar, V., Hoyle, A., Sachan, M., Ash, E.: Revisiting automated topic model evaluation with large language models. Conference on Empirical Methods in Natural Language Processing, pp. 9348-9357, Assossiation for Computational Linguistics (ACL 2023).
22. Pawlish, M., Varde, A., Robila, S.: Cloud computing for environment-friendly data centers. In proceedings of the 4th International Workshop on Cloud Data Management, pp. 43-48 (2012).
23. Pawlish, M., Varde, A.: A decision support system for green data centers. In proceedings of the 3rd Workshop on Ph.D. students in information and knowledge management, pp. 47-56 (2010).
24. Shrestha, S., Buckley, B., Varde, A., Cwynar, D.: Hybrid CNN-LSTM and Domain Modeling in Climate-Energy Analysis for a Smart Environment. IEEE 35th International Conference on Tools with Artificial Intelligence, pp. 229-233 (2023).
25. Singh, A., Yadav, J., Shrestha, S., Varde, A.: Linking alternative fuel vehicles adoption with socioeconomic status and air quality index. DOI: https://doi.org/10.48550/arXiv.2303.08286. The 37th AAAI Conference on Artificial Intelligence (2023).
26. Varde, A., Pandey, A., Du, X.: Prediction tool on fine particles pollutants and air quality for environmental engineering. SN Computer Science 3, article number 184 (2022).
27. Gonzalez-Moodie, B., Daiek, S., Lorenzo-Trueba, J., Varde, A.: Multispectral Drone Data Analysis on Coastal Dunes. IEEE International Conference on Big Data, pp. 5903-5905 (2021).
28. Prasad, A., Varde, A., Gottimukkala, R., Alo, C., Lal, P.: Analyzing Land Use Change and Climate Data to Forecast Energy Demand for a Smart Environment. 9th International Renewable and Sustainable Energy Conference, pp. 1-6 (2021).
29. Suchanek, F. M., Varde, A. S., Nayak, R., Senellart, P.: The hidden Web, XML and the semantic Web: Scientific data management perspectives. In Proceedings of the 14th International Conference on Extending Database Technology, pp. 534-537 (2011).
30. Varde, A. S., Takahashi, M., Rundensteiner, E. A., Ward, M. O., Maniruzzaman, M., Sisson Jr, R. D.: Apriori algorithm and game-of-life for predictive analysis in materials science. International Journal of Knowledge-based and Intelligent Engineering Systems, 8(4), 213-228, (2004).
31. Tandon, N., Varde, A. S., de Melo, G.: Commonsense knowledge in machine intelligence. ACM SIGMOD Record, 46(4), 49-52 (2018).